\documentclass[11pt,a4paper,twocolumn]{article}
\usepackage[a4paper,left=18mm,right=18mm,top=24mm,bottom=18mm,columnsep=7mm]{geometry}
\usepackage{tgtermes}
\usepackage{microtype}
\usepackage{graphicx}
\usepackage{booktabs}
\usepackage{tabularx}
\usepackage{array}
\usepackage{enumitem}
\usepackage{amsmath}
\usepackage{hyperref}
\usepackage{xurl}
\usepackage{balance}
\usepackage{caption}
\usepackage{titlesec}
\hypersetup{hidelinks}
\titleformat{\section}{\bfseries\large}{\thesection}{0.35em}{}
\titlespacing*{\section}{0pt}{0.7em}{0.25em}
\titleformat{\subsection}{\bfseries\normalsize}{\thesubsection}{0.35em}{}
\titlespacing*{\subsection}{0pt}{0.55em}{0.2em}
\titleformat{\subsubsection}{\bfseries\normalsize}{\thesubsubsection}{0.35em}{}
\titlespacing*{\subsubsection}{0pt}{0.45em}{0.15em}
\begin{document}
\twocolumn[{
\begin{@twocolumnfalse}
\begin{center}
{\fontsize{16}{18}\selectfont\bfseries Feature-Based Explainability of Transformer Scoring Models: A\\ Fidelity-First Framework for Automated Writing Assessment\par}
\vspace{2.0em}
{\fontsize{11.5}{13}\selectfont\bfseries Michael Hemenway\textsuperscript{1} and Edward W. Wolfe\textsuperscript{2}\par}
\vspace{0.45em}
{\fontsize{10.5}{12}\selectfont \textsuperscript{1} Pearson\\\textsuperscript{2} Iowa Testing Programs, University of Iowa\par}
\end{center}
\vspace{1.35em}
\end{@twocolumnfalse}
}]
\begin{center}\bfseries Abstract\end{center}
Transformer-based automated writing scoring models can achieve strong agreement with human scores while providing limited transparency into the characteristics of responses associated with their predictions. This study evaluated a feature-based approach to post-hoc explanation in which explanatory fidelity was established before feature importance was interpreted. English Language Arts responses from three writing prompts administered in Grades 6 and 8 were analyzed, with approximately 12,700 responses per prompt and separate DeBERTa-v3-base research models for two analytic writing traits, Expression and Conventions. Each response was represented using 61 observable linguistic and writing-quality features. Four approaches---theory-driven composites, hierarchical feature groupings, exploratory factor analysis, and Random Forest surrogates---were compared on their ability to reproduce DeBERTa predictions. Random Forest surrogates produced the largest observed $R^2$ values across all six prompt-by-trait conditions, with held-out $R^2$ values ranging from .81 to .95. SHAP analyses of these surrogates indicated substantial overlap in high-importance features across traits: 75\% to 90\% of the 20 highest-ranked features were shared. Trait differentiation occurred primarily through relative feature weighting, with Expression showing greater importance for lexical and conceptual indicators and Conventions showing greater importance for sentence-level and syntactic-control indicators. Under the researcher-defined mappings, importance-weighted rubric correspondence was substantially higher for Expression (90\%--95\%) than for Conventions (12\%--18\%). These percentages are descriptive and conditional on the mappings and available linguistic features. The findings demonstrate a fidelity-first framework for using interpretable response features to examine apparent scoring priorities while maintaining a distinction between post-hoc evidence about model behavior and direct explanation of transformer internals.\par
\noindent\textit{Keywords:} automated essay scoring, explainable artificial intelligence, transformer models, SHAP, surrogate models, validity\par
\section{Introduction}
Automated scoring of student writing has evolved from systems built primarily around human-specified linguistic features toward neural approaches that learn representations from text more directly (Ke \& Ng, 2019). Traditional automated essay scoring (AES) systems commonly relied on engineered indicators of characteristics such as vocabulary, syntax, discourse structure, and mechanics, making the relationship between the predictors and resulting scores comparatively accessible to inspection. Neural models reduced some of the dependence on explicit feature engineering by learning representations from the response itself (Nadeem et al., 2019). More recently, pretrained transformer models such as BERT have been applied to AES, extending broader developments in natural language processing to writing assessment (Mayfield \& Black, 2020). Although transformer approaches can achieve strong predictive performance, their advantage over traditional approaches is not uniform across scoring tasks (Mayfield \& Black, 2020). Moreover, unlike traditional feature-based AES models that retain direct mappings between predictors and recognizable characteristics of writing, the representations learned by neural and transformer models are substantially more difficult to characterize in those terms (Mayfield \& Black, 2020; Nadeem et al., 2019).\par
This loss of direct interpretability creates an important challenge for automated-scoring validation. Validation of an automated scoring system cannot rest solely on agreement between automated and human-assigned scores; rather, evidence is needed to support the interpretations and uses of the resulting scores and to evaluate the extent to which the scoring process reflects the intended construct (American Educational Research Association [AERA], American Psychological Association [APA], \& National Council on Measurement in Education [NCME], 2014; Williamson et al., 2012). This issue is especially salient for automated scoring because complex or proprietary scoring algorithms can make it difficult for test users to evaluate how score-producing processes relate to the qualities the assessment is intended to measure. Recent validity research in automated writing assessment similarly distinguishes evidence about human-machine score correspondence from an explanation inference concerning the relevance of features used by automated models (Hannah et al., 2023). Thus, strong agreement with human scores is important evidence about scoring performance, but it does not by itself indicate which observable qualities of writing are associated with a model's predictions.\par
Explainable artificial intelligence (XAI) methods offer one way to investigate this problem. Post-hoc explanation methods attempt to characterize the behavior of an already-trained complex model without requiring the original model to be inherently interpretable. Model-agnostic approaches can, for example, approximate the predictions of a complex model using a more interpretable surrogate or assign importance values to observable features associated with predictions (Lundberg \& Lee, 2017; Ribeiro et al., 2016). Such methods are attractive for transformer-based scoring because they can provide interpretable evidence about model behavior without requiring the transformer's learned representations to correspond directly to familiar linguistic constructs. Importantly, however, a post-hoc surrogate explains the behavior it can reproduce; it does not reveal the internal computations or reasoning process of the original model. More generally, XAI research distinguishes explanation of model behavior from direct transparency into model internals and identifies fidelity to the model being explained as an important property of surrogate explanations (Schwalbe \& Finzel, 2024).\par
Explanatory fidelity therefore constrains how strongly a surrogate can be interpreted as evidence about the target model. An interpretable surrogate may produce clear coefficients or feature rankings while reproducing the target model only imperfectly, in which case those summaries characterize the surrogate better than the target. Ribeiro et al. (2016) identify fidelity as an essential criterion for surrogate explanation and distinguish local fidelity from the more demanding problem of global fidelity. Accordingly, feature-importance interpretations are most defensible when the interpretable approximation first demonstrates substantial fidelity to the behavior being explained.\par
Feature-based explanation is particularly relevant to writing assessment because observable linguistic characteristics can potentially connect otherwise opaque score patterns to constructs that assessment developers already use to describe writing quality. Hannah et al. (2023), for example, frame the theoretical relevance of the features underlying AES scores as part of an explanation inference within an argument-based validity framework. More recent work has used textual features to examine differences in the scoring patterns of human and AI raters, including the relative weighting of grammatical accuracy, lexical sophistication, syntactic complexity, content, and other characteristics of writing (Wang et al., 2026). Such analyses can provide diagnostic evidence about whether model behavior is consistent with intended scoring dimensions and can identify patterns that warrant further investigation. At the same time, associations between linguistic features and automated scores should not be interpreted as evidence that those features constitute the transformer's internal decision process. Their value is instead as an interpretable representation against which observed score behavior can be examined.\par
The present study makes explanatory fidelity an explicit first step. Rather than selecting one feature-based method and proceeding directly to interpretation, we compare several approaches for reproducing DeBERTa predictions from the same interpretable linguistic and writing-quality features. These approaches range from theoretically and empirically derived summaries to a flexible nonlinear surrogate. The approach that most closely reproduces DeBERTa predictions then provides the basis for examining apparent scoring priorities. This ordering reflects the premise that feature importance is more informative about transformer score behavior when the explanatory model first demonstrates substantial fidelity.\par
After establishing explanatory fidelity, we examine two additional aspects of model behavior that are relevant to the interpretation of analytic writing scores. First, feature-importance patterns are compared for models trained to score Expression and Conventions. Because many characteristics of writing may be associated with both traits, the central issue is not whether each trait is associated with a completely distinct feature set, but whether the relative importance of those features differs across the independently trained trait models. Second, high-importance features are compared with dimensions represented in the trait-specific scoring rubrics. This analysis provides a structured means of identifying intended dimensions that are well represented among prominent explanatory features, expected dimensions that appear weakly represented, and important features that do not map readily onto rubric constructs.\par
Accordingly, the study addresses four research questions:\par
\begin{enumerate}[leftmargin=1.6em,itemsep=0.2em,topsep=0.2em]
\item Explanatory fidelity: How closely do alternative feature-based approaches reproduce the score predictions generated by the DeBERTa automated scoring models?
\item Apparent scoring priorities: Within the Random Forest surrogates selected on the basis of observed fidelity, which observable linguistic and writing-quality features are most strongly associated with DeBERTa score predictions?
\item Trait differentiation: How do patterns of feature importance differ between models trained to score Expression and Conventions?
\item Rubric correspondence: To what extent do the feature patterns associated with transformer scores correspond to the writing dimensions represented in the trait-specific scoring rubrics?
\end{enumerate}
\section{Method}
The analysis involved two distinct modeling stages. In the first stage, transformer-based automated scoring models were trained to predict human-assigned writing scores. In the second stage, the predictions produced by those transformer models became the outcomes for a set of feature-based explanatory analyses. The purpose of the second stage was therefore not to construct another automated scoring model of the human scores, but to determine how closely observable and interpretable characteristics of the responses could reproduce the score behavior of the trained transformer models. Alternative feature-based representations were subsequently compared on this criterion before feature importance was interpreted.\par
\subsection{Data and Scoring Context}
The study used English Language Arts writing responses from three writing prompts administered in Grades 6 and 8. Approximately 13,000 responses were available for each prompt; study documentation reports prompt-level sample sizes ranging from 12,683 to 12,970. Each response received human-assigned scores on two analytic writing traits, Expression and Conventions.\par
The prompts were treated as separate scoring contexts throughout the analysis. Responses were not pooled across prompts to fit a common transformer scoring model. Similarly, Expression and Conventions were modeled separately within each prompt. Crossing three prompts with two analytic traits therefore produced six prompt-by-trait scoring conditions.\par
This separation was retained in the explainability analyses. That is, the study did not treat the six transformer models as interchangeable instances of a single scoring function and then estimate one common explanatory model. Instead, the score behavior of each prompt-by-trait model was examined separately. Comparisons across prompts and traits were made only after the explanatory analyses had been completed within those conditions.\par
The human-assigned trait scores and the transformer predictions played different roles in the two stages of the study. Human-assigned scores served as the criterion outcomes when the DeBERTa scoring models were developed and evaluated. Once those models had been trained, however, the human scores were no longer the target of the explanation analyses. For each response, the relevant DeBERTa prediction was paired with an interpretable profile of linguistic and writing-quality features. The resulting response-level records therefore contained both the score behavior to be explained---the DeBERTa prediction---and the observable response characteristics used to construct the explanation.\par
This distinction is consequential for interpretation. Fidelity statistics from the feature-based analyses do not indicate how accurately the linguistic features predict human writing scores. They indicate how much of the variation in DeBERTa predictions can be reproduced from those features. Human-machine agreement and explanatory fidelity consequently address different questions and are reported separately.\par
\subsection{Transformer Scoring Models}
Separate DeBERTa-v3-base scoring models were developed for this research for each of the six prompt-by-trait combinations. Each model took the student's written response as input and was trained against the human-assigned score for the corresponding trait. For the research analyses, each response to a given prompt received a prediction from one model trained for Expression and a separate model trained for Conventions; models trained for one prompt were not applied to responses from another prompt. These models were developed solely for this study and were not used to assign scores for student reporting.\par
The DeBERTa models served as the target models, or explananda, for the subsequent explainability analyses. The present study was not designed to compare DeBERTa with alternative transformer architectures or to determine whether DeBERTa was the optimal automated scoring architecture. Rather, the trained models supplied a set of comparatively opaque score predictions whose observable response-level correlates were then investigated using feature-based methods.\par
Performance of the six DeBERTa research models against the human-assigned trait scores was evaluated using quadratic weighted kappa (QWK). Across prompt-by-trait conditions, QWK values ranged from approximately .78 to .89. These statistics establish the level of correspondence between the research-model predictions and human-assigned scores for the models whose behavior was subsequently examined. They are conceptually distinct from the fidelity statistics used later in the study: QWK describes agreement between DeBERTa predictions and human scores, whereas explanatory fidelity describes agreement between a feature-based surrogate and DeBERTa.\par
For the explanation stage, the response-level numeric score prediction output by the relevant trained DeBERTa model---not the human-assigned score---served as the target outcome. The same response was independently represented by the 61 interpretable linguistic and writing-quality features, and feature-based models were fit to reproduce that DeBERTa prediction.\par
The six prompt-by-trait conditions remained separate during this procedure. Explanatory fidelity was therefore estimated for each transformer model individually rather than from a single pooled analysis. This design permitted the study to determine whether a given explanatory approach performed consistently across models and provided the basis for later comparisons of feature-importance patterns between Expression and Conventions.\par
\subsection{Interpretable Feature Representation}
Each student response was represented by 61 observable linguistic and writing-quality features. This representation was constructed independently of the internal representations produced by DeBERTa and was intended to provide a parallel, human-interpretable description of the response. The feature set included relatively direct measures of response production, vocabulary, syntax, readability, cohesion, and mechanics as well as derived indicators intended to represent more complex characteristics of writing. The same feature representation was available to the explanatory analyses for both traits; features were not assigned trait-specific weights during extraction or construction.\par
Feature extraction relied primarily on the en\_core\_web\_sm English pipeline in spaCy for sentence segmentation, tokenization, lemmatization, part-of-speech tagging, dependency parsing, and noun-phrase identification. Readability indices were calculated using textstat. Additional lexical, discourse, task-alignment, and mechanics indicators were computed directly from the response text using predefined lexical lists, linguistic annotations, and rule-based procedures.\par
The initial feature set represented five broad linguistic domains together with length-normalized and task-aligned measures. Fluency and readability measures included average sentence and word length and the Flesch Reading Ease, Flesch--Kincaid Grade Level, and Coleman--Liau indices. Lexical measures included type-token ratio, lexical diversity, the proportion of unique words, the proportion of words identified as academic or comparatively sophisticated, and related vocabulary indicators. The lexical-diversity measure used a simplified MTLD-like sequential calculation with a type-token-ratio threshold of .72; responses with fewer than 10 alphabetic tokens received a value of zero for this measure.\par
Several lexical measures should be interpreted as proxies rather than direct estimates of the construct implied by their programmatic labels. For example, the variable labeled avg\_word\_frequency in the extraction code was calculated as the inverse of mean word length plus one rather than from an external frequency corpus. Similarly, the sophisticated-word measure identified words from a predefined lexical set or words longer than eight characters. The academic-word measure used a predefined academic-word lexicon, with a core fallback list available when the external list was not accessible. These measures therefore characterize observable lexical properties of the responses but should not be interpreted as comprehensive measures of lexical frequency or sophistication.\par
Syntactic features were obtained from the spaCy parses. They included mean dependency-tree depth, the estimated number of clauses per sentence, subordinate-clause ratio, average noun-phrase length, and variance in sentence length. Clause count was approximated by the number of verbs within a sentence, whereas subordinate structures were identified through dependency relations associated with markers and subordinate or adjectival clauses. Average clause length was separately approximated from distances between punctuation marks. These measures were supplemented by derived indicators such as complexity density, syntactic maturity, and controlled elaboration.\par
Cohesion and organization measures included use of transition expressions, pronoun use, lexical overlap between adjacent sentences, and causal and temporal connectives. Adjacent-sentence lexical overlap was calculated as the Jaccard overlap between sets of lemmatized alphabetic tokens in consecutive sentences. A derived cohesion\_strength indicator combined lexical overlap with the estimated number of sentences in the response. In the enhanced feature-processing stage used for the explainability analyses, this measure was log-transformed and then rescaled to the 0--1 range to reduce the disproportionate influence of its original scale. Additional derived measures represented transition density, structural consistency, organizational coherence, logical flow, and cohesion efficiency.\par
Mechanics and usage were represented using rule-based proxy measures rather than a comprehensive grammar-checking system. Punctuation problems were estimated from patterns such as repeated terminal punctuation, spaces preceding punctuation, and absent spaces following punctuation. A grammar-issue count used a limited dependency-based check for selected subject-verb agreement patterns. Word-form consistency represented the proportion of alphabetic tokens identical to their lemmatized forms, and verb-tense consistency represented the proportion of verbs belonging to the most frequently occurring verb-tag category within a response. Derived measures combined these indicators into measures such as mechanical precision, usage stability, conventions consistency, error severity, and syntactic control. Because these variables capture selected observable regularities rather than the full range of grammatical and mechanical errors, they are treated throughout the study as mechanics and usage proxies.\par
The feature set also included indicators designed to characterize task completion and higher-order aspects of the response. Predefined lexical markers were used to estimate references to textual evidence, presence of a claim or position, explanatory reasoning, and use of literary or analytical terminology. These marker counts were normalized by the number of alphabetic words in the response. Response completeness was represented as a bounded function of sentence count, increasing proportionally through five sentences and taking a maximum value of 1.0 thereafter. These measures were subsequently combined with lexical, syntactic, and cohesion measures to construct indicators such as analytical sophistication, lexical precision, interpretive depth, argument development, evidence strength, reasoning quality, expressive range, and stylistic effectiveness.\par
The derived indicators were deterministic functions of the more elementary features rather than additional information obtained from DeBERTa. For example, analytical sophistication combined vocabulary-sophistication density, academic-word use, and syntactic-complexity density; evidence strength combined text-evidence density with response completeness; and expressive range combined lexical diversity with vocabulary-sophistication density. Comparable composites were constructed for organization, style, mechanics, usage, and complexity control. Importantly, these derived variables were computed for all responses and were made available to both trait analyses rather than being differentially weighted according to the trait being modeled.\par
The final feature representation therefore contains measures that vary substantially in their proximity to the observed text. Some are direct or nearly direct characteristics of the response, such as response length or punctuation patterns; others are linguistic summaries derived from NLP annotations; and still others are composites intended to make combinations of observable characteristics more substantively interpretable. This distinction is important when interpreting subsequent feature-importance results: importance assigned to a composite such as analytical sophistication refers to that explicitly constructed combination of observable features and should not be taken to imply that DeBERTa contains an internal representation corresponding to that construct.\par
Several of the composite indicators were explicitly motivated by dimensions represented in the scoring rubrics. Consequently, the later rubric-correspondence analysis is best regarded as a descriptive examination of how explanatory importance is distributed across rubric-relevant feature representations, rather than as a statistically independent validation of the rubric-feature relationship. The feature-based representation remains external to DeBERTa in all cases; its purpose is to determine how much of the transformer's observed scoring behavior can be reproduced using response characteristics that can be described and inspected directly.\par
\subsection{Feature-Based Explanatory Approaches}
Four feature-based approaches were applied separately within each prompt-by-trait condition. In every approach, the outcome was the predicted score from the corresponding DeBERTa model rather than the human-assigned score. The approaches differed primarily in how the 61 observable features were represented before being used to reproduce the transformer prediction. Three approaches reduced the original feature space into broader scores derived theoretically or empirically, whereas the fourth modeled the individual features directly using a nonlinear Random Forest surrogate.\par
Before the feature-based analyses, responses without a target score were excluded and missing feature values were replaced with the corresponding feature median. Features were standardized to a mean of zero and standard deviation of one. This standardized representation was used for the theory-driven, hierarchical, and EFA approaches and was also supplied to the Random Forest surrogate.\par
\subsubsection{Theory-Driven Representation}
The theory-driven approach organized the observed features into six predefined writing-related groups: Fluency/Readability, Lexical Sophistication, Syntactic Complexity, Cohesion, Grammar/Mechanics, and Task Alignment. Group membership was specified prior to fitting the explanatory models rather than determined from the empirical relationships between features and DeBERTa predictions.\par
A score for each group was calculated by averaging the standardized values of the features assigned to that group. The resulting six scores therefore represented equal-weight composites of the available features rather than estimated latent-variable scores. Some features contributed to more than one theoretical group when they were judged relevant to multiple aspects of writing. For example, response completeness contributed to both Fluency/Readability and Task Alignment, and literary-language use contributed to both Lexical Sophistication and Task Alignment.\par
The six composite scores were entered simultaneously into an ordinary least-squares regression model predicting the corresponding DeBERTa score. Regression coefficients described the conditional association of each theory-defined group with the transformer prediction after accounting for the remaining group scores. Explanatory fidelity for this approach was summarized by the proportion of variation in DeBERTa predictions reproduced by the regression model.\par
\subsubsection{Hierarchical Feature Representation}
The hierarchical approach replaced the predefined feature groups with empirically derived clusters of related features. Pairwise correlations were first calculated among the standardized features. Feature dissimilarity was defined as $1 - |r|$, so strongly positively or negatively correlated features were treated as similar for purposes of clustering. Hierarchical agglomerative clustering was then performed using Ward linkage.\par
The number of feature clusters was not fixed in advance. Solutions containing three through ten clusters were evaluated. For each candidate solution, the standardized features within each cluster were averaged to form a cluster score, and those scores were entered into a linear regression predicting the corresponding DeBERTa prediction. The number of clusters was selected using an elbow procedure based on changes in $R^2$ across successive solutions. Specifically, the procedure identified the point showing the largest decrease in the rate of improvement in $R^2$.\par
After the number of clusters was selected, the final cluster scores were computed as the mean of the standardized features assigned to each cluster and entered simultaneously into an ordinary least-squares regression model. Feature membership in the clusters was therefore determined from interfeature relationships, whereas the choice of the number of clusters incorporated information about how well the resulting representation reproduced DeBERTa scores.\par
\subsubsection{Exploratory Factor Analysis}
The third reduced-dimension representation was based on exploratory factor analysis (EFA). Before factor extraction, the standardized features were screened for extreme multicollinearity. When two features correlated at an absolute value greater than .90, one member of the pair was removed, with preference given to retaining the feature having the lower average absolute correlation with the remainder of the feature set.\par
The number of factors was selected using parallel analysis. Eigenvalues from the observed feature correlation matrix were compared with eigenvalues generated from 100 random normal datasets of the same dimensions. A factor was retained when its observed eigenvalue exceeded the 95th percentile of the corresponding random-data eigenvalues. The number of retained factors was capped at eight to preserve interpretability.\par
The retained factors were estimated using the FactorAnalysis procedure with a maximum of 1,000 iterations, a random-state value of 42, and varimax rotation. Factor scores were then calculated for each response. Although an earlier study proposal described an oblique factor rotation, the archived analysis implementation used varimax rotation; the procedure described here reflects the implementation used in the analysis pipeline.\par
The resulting factor scores were entered simultaneously into an ordinary least-squares regression predicting the corresponding DeBERTa score. An optional clustering stage was available in the EFA software, but it was disabled in the multi-trait analyses reported here. Thus, the EFA approach evaluated whether a relatively small set of empirically estimated latent dimensions could reproduce transformer score behavior.\par
\subsubsection{Random Forest Surrogate and SHAP}
The fourth approach retained the individual linguistic and writing-quality features rather than first aggregating them into theory-defined groups, empirical clusters, or factors. A Random Forest regression model was trained to predict the corresponding DeBERTa score from the feature set. The data were divided randomly into 80\% training and 20\% test samples using a random-state value of 42. The Random Forest contained 100 trees, used a maximum tree depth of 10, and otherwise allowed trees to use all available processor cores during fitting.\par
The Random Forest therefore functioned as a global surrogate for the DeBERTa scoring model: it learned a flexible nonlinear mapping between the observable response features and the transformer predictions. Unlike the three reduced-dimension approaches, it did not require the effects of features to be linear or additive and could represent interactions and threshold relationships among predictors.\par
Surrogate fidelity was evaluated on the held-out 20\% test sample using $R^2$. SHAP values were then calculated for the Random Forest using TreeExplainer (Lundberg \& Lee, 2017). The explainer used the interventional feature-perturbation formulation and a background sample of up to 500 randomly selected training responses. SHAP values were calculated for responses in the test sample.\par
Global feature importance was defined as the mean absolute SHAP value for each feature across the test responses. This quantity represents the average magnitude of the feature's contribution to predictions from the Random Forest surrogate without regard to whether the contribution increased or decreased an individual prediction. Feature importance was also summarized within the theory-defined feature families for descriptive purposes, but subsequent analyses of apparent scoring priorities and trait differences focused primarily on the individual-feature SHAP results.\par
The SHAP values directly explain predictions from the Random Forest rather than from DeBERTa itself. Their relevance to the transformer therefore depends on the extent to which the Random Forest reproduces the DeBERTa predictions. For that reason, SHAP feature importance was not interpreted until the fidelity of the Random Forest surrogate had been compared with the alternative feature-based approaches.\par
\subsection{Evaluation of Explanatory Fidelity}
Explanatory fidelity was evaluated separately for each of the six prompt-by-trait scoring conditions. For each condition, the predictions from the relevant DeBERTa model served as the reference values, and predictions from each feature-based explanatory model were compared with those reference values. The primary fidelity statistic was the coefficient of determination, $R^2$, defined as\par
\begin{equation*} R^2 = 1 - \frac{\sum_i (y_i-\hat{y}_i)^2}{\sum_i (y_i-\bar{y})^2} \end{equation*}
where $y_i$ is the DeBERTa prediction for response i, $\hat{y}_i$ is the corresponding prediction from the feature-based explanatory model, and $\bar{y}$ is the mean DeBERTa prediction. In this context, $R^2$ represents the proportion of variation in DeBERTa score predictions reproduced by the explanatory model. Higher values therefore indicate greater fidelity to the score behavior being explained rather than greater agreement with the human-assigned scores.\par
The theory-driven, hierarchical, and EFA representations were each evaluated using the predictions from the ordinary least-squares regression fitted to their respective composite, cluster, or factor scores. For these three approaches, $R^2$ was calculated from fitted values for the same responses used to estimate the regression model. Root mean squared error and mean absolute error were also calculated by the analysis software as descriptive model-fit indices, although $R^2$ was used as the common criterion for comparing explanatory fidelity across approaches.\par
The hierarchical procedure incorporated $R^2$ at an earlier stage as well. Candidate solutions containing three through ten feature clusters were each converted to cluster scores and used to predict DeBERTa scores. The observed change in $R^2$ across those solutions contributed to selection of the final number of clusters. The fidelity value subsequently reported for the hierarchical approach therefore reflects the selected cluster solution rather than a cluster count fixed independently of the target predictions.\par
The Random Forest approach used a different evaluation design. Its surrogate was estimated on the randomly selected 80\% training partition, and fidelity was calculated from predictions for the held-out 20\% test partition. Both training and test $R^2$ were generated during model fitting, but the test-set $R^2$ was used in the cross-approach fidelity comparison and in the reported results. The SHAP analyses likewise used responses from this held-out test partition.\par
This difference in evaluation design is important for interpreting the fidelity comparison. The $R^2$ values for the three reduced-dimension approaches describe in-sample reproduction of DeBERTa predictions, whereas the Random Forest value describes reproduction of predictions for held-out responses. The four values therefore should be interpreted as descriptive indicators of fidelity rather than as strictly equivalent estimates of out-of-sample performance. No resampling procedure was used to place all four approaches on a common held-out evaluation framework in the analyses reported here.\par
The fidelity comparison was conducted separately within each prompt-by-trait condition, yielding four $R^2$ values for each DeBERTa model. The purpose was to identify the feature-based representation that produced the largest and most consistent observed $R^2$ values across conditions. Feature importance from the selected approach was then carried forward for interpretation of apparent scoring priorities, trait differentiation, and rubric correspondence.\par
\subsection{Feature-Importance Analysis}
Feature-importance analyses were conducted after the fidelity comparison and were based on the Random Forest surrogates. For each prompt-by-trait condition, individual-feature importance was defined as the mean absolute SHAP value across responses in the Random Forest test sample. Features were ranked from largest to smallest mean absolute SHAP value. Because absolute SHAP values were used, these rankings describe the magnitude of a feature's contribution to the surrogate predictions rather than the direction of its association with predicted score.\par
The first analysis characterized the extent to which the two trait models within a prompt emphasized a common set of features. For each trait, the 20 features with the largest mean absolute SHAP values were identified. The intersection of the two top-20 sets was then calculated, together with the features appearing among the top 20 for only one trait. Cross-trait overlap was expressed as\par
\begin{equation*}\mathrm{Overlap}=\frac{\text{number of shared top-20 features}}{20}\times100\end{equation*}
The overlap analysis therefore addressed whether the Expression and Conventions surrogates drew heavily on the same observable response characteristics, without assuming that a shared feature carried the same amount of explanatory importance for both traits.\par
A second analysis examined differences in the relative importance of individual features across the two trait models. Mean absolute SHAP importance values from the Expression and Conventions surrogates were aligned by feature. For feature j, the absolute difference in importance was calculated as\par
\begin{equation*} D_j = |I^E_j-I^C_j| \end{equation*}
where $I^E_j$ and $I^C_j$ are the mean absolute SHAP importance values for Expression and Conventions, respectively. Features were ranked by $D_j$, and the 25 features showing the largest absolute differences were retained for the differential comparison.\par
A signed relative-difference index was also calculated as\par
\begin{equation*} R_j = \frac{I^E_j-I^C_j}{\max(I^E_j,I^C_j)} \end{equation*}
with the value set to zero when both importance values were zero. Values near zero indicate similar importance across the two trait surrogates, whereas values farther from zero indicate greater differentiation. The sign indicates which trait received the larger mean absolute SHAP importance. An importance ratio, calculated as ($I^E_j + 10^{-6}$)/($I^C_j + 10^{-6}$), was retained as an additional descriptive comparison and avoided division by zero through the small additive constant.\par
These cross-trait calculations used the mean absolute SHAP importance values produced separately by the two Random Forest surrogates; the importance values were not additionally rescaled or normalized across traits before the comparison. The analyses should therefore be interpreted as descriptive contrasts between the explanatory profiles of the two independently estimated surrogates rather than as formal tests of equality in feature effects.\par
The overlap and differential analyses address related but distinct questions. A feature may rank highly for both traits and therefore contribute to substantial top-feature overlap while nevertheless carrying appreciably more explanatory weight for one trait. Conversely, a large proportional difference may occur for a feature whose absolute importance is small in both models. Differential results were therefore interpreted jointly with overall feature importance rather than from the relative-difference statistic alone.\par
Finally, these comparisons concern surrogate feature importance, not direct feature weights within DeBERTa. A finding that a feature received greater SHAP importance for one trait indicates that the feature contributed more strongly to reproducing the predictions of that trait's Random Forest surrogate. The strength with which that pattern can be attributed to DeBERTa score behavior depends on the previously established fidelity of the corresponding surrogate.\par
\subsection{Rubric-Correspondence Analysis}
The final analysis examined whether the observable features receiving the greatest explanatory importance corresponded to dimensions represented in the scoring rubrics. The analysis was conducted separately for Expression and Conventions and used the feature-importance results from the Random Forest surrogates. Its purpose was not to establish independent construct validity from the same features used in the surrogate analyses, but to provide a structured description of the extent to which prominent explanatory features could be connected to intended scoring dimensions.\par
Feature-to-rubric mappings were specified explicitly before the alignment calculations. For Expression, the mappings organized features into four rubric-related dimensions: comprehension/analysis, development/evidence, organization/coherence, and style/clarity. For Conventions, the mappings represented grammar/mechanics, usage consistency, and complexity control. Each dimension was associated with a predefined set of observable or composite features judged to represent that aspect of the rubric. A feature could appear in more than one mapping when it was considered relevant to multiple dimensions.\par
These mappings were developed by a single researcher using substantive judgment rather than statistically estimated from the scoring data, and they were not independently validated. They linked rubric language to observable feature representations and therefore embodied substantive assumptions about which linguistic indicators could reasonably provide evidence about a rubric dimension. The mappings were not generated from the SHAP rankings themselves and were not automatically altered according to the observed feature importance.\par
For each prompt-by-trait analysis, the 15 features with the largest mean absolute SHAP values were selected. Each of these features was then classified as aligned if it appeared in at least one of the predefined mappings for the relevant trait. Features that did not appear in any mapping were retained as high-importance but unmapped features. Conversely, mapped features that were not among the top 15 were identified as expected features that did not receive sufficiently high importance to enter the analysis set.\par
An overall rubric-alignment score was calculated from the SHAP importance associated with mapped features:\par
\begin{equation*} A=\frac{\sum_{j\in M}I_j}{\sum_{j\in T}I_j} \end{equation*}
where T denotes the 15 highest-importance features, M is the subset of those features represented in at least one rubric mapping, and $I_j$ is the mean absolute SHAP importance of feature j. Thus, the alignment score represents the proportion of the total importance contained in the 15 leading explanatory features that could be associated with a predefined rubric dimension. It is importance-weighted rather than simply the percentage of top features that were mapped.\par
Correspondence was also summarized separately for each rubric dimension. Dimension coverage was calculated as the number of mapped features appearing among the top 15 divided by the total number of features predefined for that dimension. The total SHAP importance contributed by the matched features was also retained. These summaries allowed a trait to show a relatively high overall alignment score while still revealing particular rubric dimensions that were weakly represented among its most important explanatory features.\par
The analysis pipeline additionally compared EFA factors with the same rubric mappings. For each retained factor, the ten features with the largest absolute loadings were identified and their mappings to rubric dimensions were counted. The rubric dimension receiving the largest number of matches was designated the factor's primary correspondence, and the proportion of the ten leading loading features assigned to that dimension was retained as a descriptive correspondence index. Factor regression coefficients were used to characterize the relative importance of the factors. These factor-level results provided supplementary context but did not determine the principal alignment scores reported for the Random Forest/SHAP analyses.\par
A large language model, GPT-5-mini, was used after the quantitative mapping calculations to assist with semantic interpretation of the resulting correspondence patterns. The model received the trait-specific rubric text, the calculated feature-alignment results, information about unmapped high-importance and absent expected features, and, when available, the factor-level correspondence results. It generated narrative summaries of apparent strengths, misalignments, rubric gaps, and interpretive considerations. The LLM did not compute SHAP values, determine feature importance, create the predefined feature-to-rubric mappings, or calculate the alignment scores. Quantitative correspondence results were produced programmatically from the predefined mappings and observed feature-importance values.\par
The rubric-correspondence analysis therefore combines a reproducible quantitative procedure with an explicitly judgment-based mapping between linguistic features and rubric constructs. The resulting alignment percentages should be interpreted as descriptive indices conditional on the researcher-defined mappings and available linguistic features, rather than as objective measures of construct alignment or direct measures of construct validity. In particular, weak correspondence can reflect limitations in the observable feature set or in the feature-to-rubric mapping as well as differences between apparent model behavior and rubric emphasis.\par
\section{Results}
Results are organized around the four research questions. The analysis first compared the observed explanatory fidelity of the four feature-based approaches. Because Random Forest surrogates produced the largest $R^2$ values across all six prompt-by-trait conditions, subsequent results use their SHAP values to characterize apparent scoring priorities, compare the two analytic traits, and examine rubric correspondence.\par
\subsection{Explanatory Fidelity}
The four feature-based approaches differed substantially in the amount of variation in DeBERTa predictions that they reproduced. Table 1 summarizes the observed $R^2$ ranges across the six prompt-by-trait models. The Random Forest surrogate produced the largest observed $R^2$ in every scoring condition.\par
\begin{table}[t]
\centering
\caption{\textit{Explanatory Fidelity Across Prompt-by-Trait Conditions}}
\vspace{-0.4em}
\small
\begin{tabularx}{\columnwidth}{X>{\raggedleft\arraybackslash}p{0.25\columnwidth}}
\toprule
\textbf{Approach} & \textbf{$R^2$ range} \\
\midrule
Theory-driven & .33--.57 \\
Hierarchical & .26--.55 \\
Exploratory factor analysis & .43--.86 \\
Random Forest surrogate & .81--.95 \\
\bottomrule
\end{tabularx}
\vspace{0.15em}
\begin{minipage}{\columnwidth}\footnotesize\textit{Note.} Random Forest values are held-out test-set $R^2$ values; the other three approaches were evaluated in sample.\end{minipage}
\end{table}
The theory-driven approach reproduced between 33\% and 57\% of the variation in DeBERTa scores across conditions. The hierarchical representation yielded a similar but somewhat lower range, from .26 to .55. EFA showed greater variability, with $R^2$ values ranging from .43 to .86. Thus, the dimension-reduction approaches could reproduce meaningful portions of transformer score variation, but their fidelity was inconsistent across scoring conditions.\par
Random Forest surrogates produced substantially larger $R^2$ values. Held-out test $R^2$ ranged from .81 to .95, indicating that the observable feature set reproduced approximately 81\% to 95\% of the variance in DeBERTa predictions for held-out responses. These were the largest observed $R^2$ values in each of the six prompt-by-trait conditions.\par
The evaluation procedures were not identical: theory-driven, hierarchical, and EFA $R^2$ values were calculated in sample, whereas Random Forest results were based on held-out responses. The values are therefore descriptive fidelity indicators rather than strictly equivalent estimates of out-of-sample performance. Nevertheless, the Random Forest produced the largest $R^2$ in every scoring condition despite the more demanding evaluation and was selected for the remaining analyses.\par
Accordingly, the SHAP values below describe the Random Forest surrogates and are interpreted as evidence about the substantial portion of DeBERTa score variation those surrogates reproduce, not as direct measurements of DeBERTa internals.\par
\subsection{Apparent Scoring Priorities}
The SHAP analyses showed several recurring features across the six Random Forest surrogates. Although the exact ordering of features varied across prompts and traits, the strongest recurring pattern involved measures related to cohesion and response completeness.\par
cohesion\_strength was the leading feature in every prompt-by-trait analysis. Because this feature combined adjacent-sentence lexical overlap with the extent of response development, its prominence indicates that a broad measure of connectedness and elaboration was consistently associated with the score patterns reproduced by the surrogates. response\_completeness also appeared consistently among the highest-importance features. cohesion\_efficiency, another derived measure combining cohesion with response length or development, was frequently prominent. Lexical overlap between adjacent sentences was also an important shared signal in one of the prompts.\par
These recurring signals combined aspects of textual development, connectedness, and response production. Because several were derived measures, their SHAP importance should be interpreted as evidence that they predict surrogate score behavior, not that DeBERTa explicitly computes corresponding constructs.\par
Other feature types contributed more selectively. Measures representing vocabulary, analytical sophistication, expressive range, readability, clause structure, and sentence control became especially informative when the two analytic traits were compared. The distinction between the traits therefore emerged less from wholly different sets of important features than from differences in the amount of explanatory importance assigned to features within a largely shared response representation.\par
\subsection{Trait Differentiation}
The two trait models showed substantial overlap in their highest-importance features. Table 2 shows the number and percentage of shared features among the top 20 SHAP-ranked features for Expression and Conventions within each prompt. Across prompts, 75\% to 90\% of the top-ranked features were shared.\par
\begin{table}[t]
\centering
\caption{\textit{Overlap in the 20 Highest-Importance Features Across Traits}}
\vspace{-0.4em}
\small
\begin{tabularx}{\columnwidth}{>{\raggedright\arraybackslash}p{0.20\columnwidth}>{\raggedright\arraybackslash}p{0.37\columnwidth}>{\raggedright\arraybackslash}p{0.33\columnwidth}}
\toprule
\textbf{Prompt} & \textbf{Shared features} & \textbf{Percentage shared} \\
\midrule
1 & 17 of 20 & 85\% \\
2 & 15 of 20 & 75\% \\
3 & 18 of 20 & 90\% \\
\bottomrule
\end{tabularx}
\vspace{0.15em}
\begin{minipage}{\columnwidth}\footnotesize\textit{Note.} Percentages are based on the 20 highest-ranked SHAP features within each trait model.\end{minipage}
\end{table}
The overlap was substantial in every prompt. This result indicates that trait differentiation did not arise primarily because each model relied on a distinct set of observable writing characteristics.\par
Instead, the clearer distinction appeared in the relative importance assigned to shared or broadly available features. Expression models tended to assign relatively greater explanatory importance to lexical precision, analytical sophistication, expressive range, and related indicators of vocabulary quality and conceptual depth. Conventions models showed relatively greater importance for average clause length, readability measures, and indicators of sentence-level and syntactic control.\par
\begin{table}[t]
\centering
\caption{\textit{Summary of Shared and Trait-Differentiating Feature Patterns}}
\vspace{-0.4em}
\small
\begin{tabularx}{\columnwidth}{>{\raggedright\arraybackslash}p{0.36\columnwidth}X}
\toprule
\textbf{Pattern} & \textbf{Features or feature types} \\
\midrule
Recurring across both traits & Cohesion strength; response completeness; cohesion efficiency; lexical overlap \\
Relatively greater importance for Expression & Lexical precision; analytical sophistication; expressive range; vocabulary and conceptual-depth indicators \\
Relatively greater importance for Conventions & Average clause length; readability; sentence-control indicators; syntactic-control indicators \\
\bottomrule
\end{tabularx}
\vspace{0.15em}
\begin{minipage}{\columnwidth}\footnotesize\textit{Note.} These categories describe relative SHAP importance patterns. They do not indicate that a feature was used exclusively for one trait.\end{minipage}
\end{table}
Figure 1 shows the differential SHAP analysis for a representative prompt. The upper panels display the relative magnitudes of feature importance and absolute between-trait differences; the lower panels express the same contrast as a signed relative difference and importance ratio.\par
\begin{figure*}[t]
\centering
\includegraphics[width=0.84\textwidth]{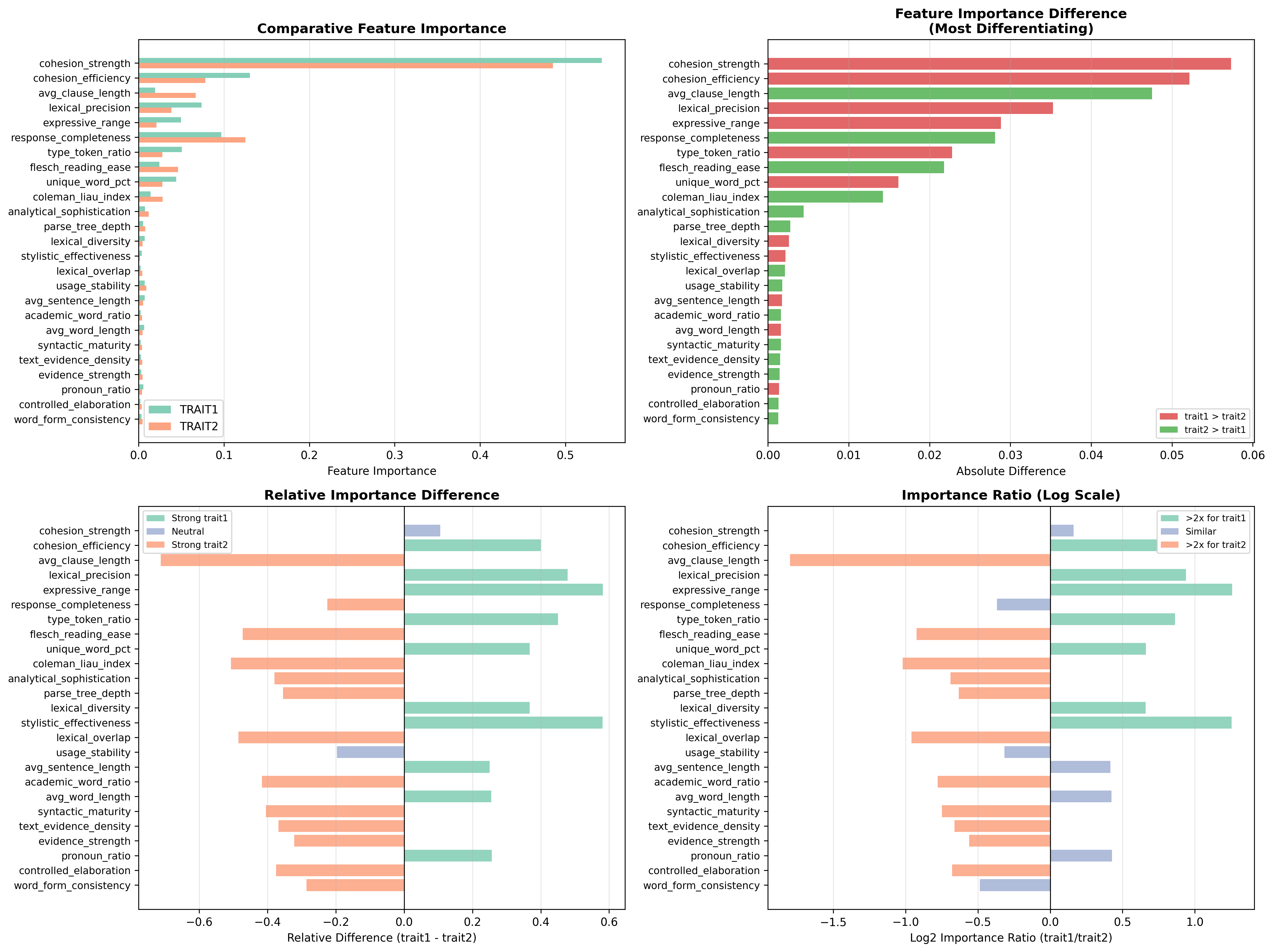}
\caption{\textit{Representative Prompt-Level Comparison of Feature Importance Across Traits}}
\begin{minipage}{0.88\textwidth}\footnotesize\textit{Note.} The original analysis output labels the two models Trait 1 and Trait 2; in the study materials these correspond to Expression and Conventions, respectively. Features farther from zero in the lower-left panel show greater differentiation in relative importance. Interpretation should also consider absolute importance because a large proportional difference can occur for a feature with little overall importance.\end{minipage}
\end{figure*}
The contrast should not be interpreted as a categorical separation of features by trait. Lexical precision, for example, also contributed to Conventions models, and syntactic measures also contributed to Expression models. The distinguishing pattern was one of relative weighting within a substantially overlapping feature space. Features near zero on the differential-importance measure showed similar importance for both traits, whereas larger positive or negative differences identified features whose explanatory contribution was more strongly associated with one trait than the other.\par
The high degree of top-feature overlap and the simultaneous presence of systematic relative differences therefore provide complementary results. Both traits appeared to draw on broad indicators of overall response quality, especially cohesion and completeness, while the additional explanatory importance of lexical and conceptual indicators for Expression and sentence-level control indicators for Conventions differentiated the two score patterns.\par
\subsection{Rubric Correspondence}
Under the researcher-defined mappings, rubric correspondence differed sharply between the two traits. The importance-weighted alignment scores for Expression ranged from approximately 90\% to 95\% across prompts. For Conventions, the corresponding values ranged from approximately 12\% to 18\%.\par
\begin{table}[t]
\centering
\caption{\textit{Rubric Correspondence of High-Importance Features}}
\vspace{-0.4em}
\small
\begin{tabularx}{\columnwidth}{>{\raggedright\arraybackslash}p{0.20\columnwidth}>{\raggedright\arraybackslash}p{0.22\columnwidth}X}
\toprule
\textbf{Trait} & \textbf{Alignment range} & \textbf{Dominant correspondence pattern} \\
\midrule
Expression & 90\%--95\% & High-importance features corresponded closely to development, organization, and style-related dimensions \\
Conventions & 12\%--18\% & Broad cohesion and completeness signals remained prominent; grammar and mechanics features were less well represented \\
\bottomrule
\end{tabularx}
\vspace{0.15em}
\begin{minipage}{\columnwidth}\footnotesize\textit{Note.} Alignment is conditional on the researcher-defined feature-to-rubric mappings and available feature set.\end{minipage}
\end{table}
For Expression, most of the SHAP importance represented among the 15 leading features could be associated with one or more predefined rubric dimensions. High-importance features such as lexical precision, cohesion strength, and expressive range corresponded to rubric language concerning development, organization, sophisticated word choice, and effective style. Thus, the observable features that most strongly reproduced Expression score behavior generally fell within areas represented in the trait rubric.\par
The Expression results did not imply uniform coverage of all intended dimensions. Because the overall alignment statistic was importance-weighted, a relatively small number of strongly weighted rubric-relevant features could contribute substantially to the total alignment percentage. Dimension-level coverage therefore remained useful for identifying rubric components for which fewer expected features appeared among the leading SHAP results. Nevertheless, at the aggregate level, the explanatory feature patterns for Expression showed strong correspondence with the predefined rubric mappings.\par
The pattern for Conventions was substantially different. Within the researcher-defined Conventions mappings, only 12\% to 18\% of the importance represented among the leading features was associated with the mapped dimensions. Broad coherence and response-completeness measures remained prominent, whereas features intended to represent grammar, mechanics, and usage were much less prominent among the highest-importance features. Sentence-level measures such as clause length, readability, and syntactic control contributed to distinguishing Conventions from Expression, but the overall SHAP profile did not map strongly onto the explicit grammar- and mechanics-oriented feature representation.\par
The low Conventions alignment scores identify a discrepancy between the strongest feature-based explanation of the research model's score behavior and the dimensions represented in the researcher-defined rubric mapping. They do not, by themselves, establish that the Conventions research model measures an inappropriate construct. Three sources of the discrepancy remain observationally indistinguishable in the present analysis: the available mechanics and usage features may provide an incomplete representation of the rubric construct; the researcher-defined feature-to-rubric mappings may omit legitimate relationships; or broader response characteristics such as cohesion and completeness may genuinely be associated with DeBERTa predictions for Conventions.\par
Taken together, the Random Forest surrogates produced the largest observed $R^2$ values; cohesion and response-completeness measures were prominent across traits and prompts; the traits shared most high-importance features but differed in relative weighting; and rubric correspondence under the researcher-defined mappings was strong for Expression but weak for Conventions.\par
\section{Discussion}
This study evaluated a fidelity-first approach to feature-based explanation of transformer scoring models: substantive interpretation followed evidence that the explanatory model reproduced the behavior being explained. Across three prompts and two analytic traits, Random Forest surrogates produced held-out $R^2$ values of .81 to .95; Expression and Conventions shared most high-importance features but differed in relative weighting; and, under the researcher-defined mappings, rubric correspondence was strong for Expression but weak for Conventions. These results support the usefulness of the approach while also illustrating limits on what post-hoc feature explanations can establish.\par
\subsection{Explanatory Fidelity as a Prerequisite for Interpretation}
All four approaches used the same observable response characteristics, but the amount of DeBERTa score variation they reproduced differed considerably. Theory-driven and hierarchical summaries produced moderate $R^2$ values, EFA varied across conditions, and the Random Forest surrogate produced the largest $R^2$ in every condition.\par
This result reinforces the distinction between an explanation that is intuitively interpretable and one that is demonstrably faithful to the model being explained. Ribeiro et al. (2016) identify fidelity as a central property of surrogate explanations, and Schwalbe and Finzel (2024) similarly distinguish post-hoc approximations of model behavior from transparency into the underlying model itself. The present results show why that distinction matters in an automated-scoring context. Had the theory-driven representation been selected solely because its categories were easy to interpret, subsequent conclusions would have characterized an explanatory representation that reproduced considerably less of the target score behavior.\par
The larger $R^2$ values observed for the Random Forest surrogate suggest that the relationship between observable features and DeBERTa predictions was not adequately represented by small linear composites. The analyses cannot distinguish whether this advantage reflects nonlinearity, interactions, or information lost through aggregation; nevertheless, they show that representational convenience can come at a cost in fidelity.\par
At the same time, the Random Forest surrogates were not perfectly faithful. An $R^2$ of .81 to .95 leaves between approximately 5\% and 19\% of DeBERTa prediction variance unreproduced by the observed feature representation and surrogate model. SHAP patterns should therefore be interpreted as characterizing the substantial portion of score behavior captured by the surrogate rather than as a complete decomposition of the transformer prediction.\par
\subsection{Shared Feature Space and Trait Differentiation}
The feature-importance results complicate a simple view in which analytically scored writing traits depend on sharply distinct sets of response characteristics. Expression and Conventions shared between 75\% and 90\% of their 20 most important explanatory features within prompts. Cohesion strength and response completeness were especially prominent across models, and related measures of cohesion efficiency and lexical overlap also appeared repeatedly.\par
The overlap is not necessarily problematic. Analytic writing dimensions are distinguishable constructs, but they are expressed through the same written response and need not be statistically independent. A well-developed response can simultaneously exhibit greater length, stronger cohesion, greater lexical sophistication, and more controlled sentence structure. Feature-based explanation should therefore not require each trait model to operate on a unique vocabulary of observable response characteristics.\par
More informative differentiation emerged in the relative weighting of those characteristics. Expression placed comparatively greater explanatory weight on lexical precision, analytical sophistication, expressive range, and related vocabulary and conceptual-depth indicators. Conventions placed comparatively greater weight on clause length, readability, and sentence- and syntactic-control measures. This pattern is more consistent with graded trait differentiation than with a partition in which some features belong exclusively to one trait.\par
This emphasis on relative weighting also parallels the broader motivation for feature-based analysis in automated scoring. Wang et al. (2026), for example, examined differences in the weighting of observable textual characteristics across scoring systems rather than treating the presence of a feature as evidence of a unique construct representation. In the present study, the important distinction is likewise not simply whether a feature was associated with a trait prediction, but how strongly the feature contributed to reproducing one trait's score pattern relative to the other.\par
\subsection{Rubric Correspondence and Construct Interpretation}
The rubric-correspondence results provide the most direct connection between the explanatory feature patterns and intended scoring constructs, but they also require the most cautious interpretation. Under the researcher-defined mappings, 90\% to 95\% of the importance represented among the leading Expression features could be mapped to the predefined rubric dimensions. The prominence of lexical precision, cohesion, expressive range, and related measures was broadly consistent with rubric emphases involving development, organization, analysis, and style.\par
This pattern is consistent with the explanation inference discussed by Hannah et al. (2023): observable characteristics associated with automated scores can contribute evidence about whether scoring behavior bears a plausible relationship to theoretically relevant writing qualities. The present approach adds an additional requirement to that logic by conditioning such interpretation on the fidelity of the explanatory model.\par
The Conventions results were much less straightforward. Under the researcher-defined Conventions mappings, only 12\% to 18\% of high-feature importance was captured by the mapped dimensions. Grammar- and mechanics-related proxy measures were not among the dominant explanatory features to the degree that would be expected from a direct reading of the rubric, whereas cohesion, completeness, and broader response-quality indicators remained prominent.\par
This discrepancy is diagnostically useful, but its source cannot be determined from the present analysis alone. One possibility is that the DeBERTa Conventions research models relied partly on broad characteristics of writing that are correlated with conventional control, even when those characteristics are not explicit components of the rubric. A second possibility is that the feature representation was inadequate for the trait. The mechanics and usage indicators were relatively simple rule-based proxies, and the ability of the surrogate to express a rubric construct is necessarily constrained by the available feature set. A third possibility is that the researcher-defined feature-to-rubric mappings were incomplete.\par
For these reasons, the low Conventions correspondence under the researcher-defined mappings should not be treated as evidence that the research scoring models are invalid. It is better understood as a descriptive finding that warrants further investigation of the Conventions research models, the available grammar and mechanics feature representation, and the mappings themselves. This diagnostic role illustrates how feature-based explainability may contribute to automated-scoring validation research.\par
\subsection{Implications for Automated-Scoring Validation}
Traditional evaluation of automated scoring appropriately emphasizes correspondence with human scores, distributional behavior, subgroup performance, reliability, and other evidence relevant to score interpretation and use (Williamson et al., 2012). Feature-based explainability addresses a different question: whether observable characteristics of responses can provide a credible account of the score behavior generated by an otherwise difficult-to-interpret model.\par
Feature-based explainability is useful only insofar as the explanatory representation reproduces the target model and conclusions remain at the level of model behavior rather than internal reasoning. A high-fidelity surrogate therefore supports stronger claims about apparent scoring priorities than an inadequately faithful one, while still remaining a surrogate explanation.\par
The fidelity-first sequence therefore provides a practical validation workflow: construct one or more interpretable representations, evaluate their ability to reproduce the target model, select an explanatory model only after considering that fidelity, and then examine feature importance, trait differentiation, and correspondence with intended constructs. This ordering prevents the availability of an attractive explanation from being mistaken for evidence that the explanation adequately represents the scoring model.\par
\section{Limitations}
Several limitations qualify the conclusions of the study.\par
First, the analyses are post-hoc and model-agnostic. The selected Random Forest surrogates do not provide direct access to DeBERTa's hidden representations, attention patterns, or computational process. The study therefore supports conclusions about observable characteristics associated with reproduced score behavior, not about the literal information used internally by the transformer.\par
Second, the four explanatory approaches were not evaluated under identical validation designs. The theory-driven, hierarchical, and EFA models were evaluated using in-sample $R^2$, whereas the Random Forest surrogate was evaluated on a held-out test partition. This difference does not weaken the observation that the Random Forest achieved the largest $R^2$ despite the more demanding evaluation, but it prevents the four statistics from serving as strictly comparable estimates of generalization performance. A future comparison should evaluate all approaches on common held-out partitions or within the same cross-validation design.\par
Third, the explanation is necessarily constrained by the feature set. The 61 features represent a substantial range of observable writing characteristics, but they are not exhaustive. Some features are relatively direct measurements, whereas others are heuristic proxies or composites. In particular, several grammar, mechanics, lexical-sophistication, and task-alignment measures use simplified operational definitions. A transformer may exploit textual characteristics for which no adequate feature was included. Consequently, absence or low importance of a feature cannot be interpreted as evidence that the corresponding writing quality is absent from transformer score behavior.\par
Fourth, the feature set contains correlated and derived variables. Composite indicators can improve interpretability, but they also redistribute information already represented in their component variables. SHAP importance in the presence of correlated predictors is therefore not equivalent to a unique causal decomposition of prediction variance. The importance rankings identify features that are useful within the fitted surrogate and the available predictor set.\par
Fifth, rubric correspondence depends on human judgment. The feature-to-rubric mappings were specified by a single researcher and were not independently validated, and the same mappings determined which feature importance counted as aligned. The alignment percentages are therefore descriptive indices conditional on the researcher-defined mappings and available linguistic features rather than objective measures of construct alignment or independent tests of construct validity. The use of a large language model to summarize alignment patterns does not remove this limitation; the quantitative mappings and alignment calculations preceded the LLM interpretation.\par
Finally, the study examined three writing prompts, two grade levels, two analytic traits, and one transformer architecture. The consistency observed across the six models provides useful internal replication, but it does not establish that the same fidelity or feature-importance patterns would occur for other prompts, grade ranges, writing constructs, assessment programs, or transformer architectures.\par
\section{Future Research}
The most immediate extension is methodological. All explanatory approaches should be evaluated under a common out-of-sample framework so that fidelity differences can be compared directly. Repeated cross-validation or bootstrap procedures could also be used to examine the stability of fidelity and feature-importance rankings across samples.\par
A second priority is improving the interpretable feature representation, particularly for Conventions. More direct measures of grammatical accuracy, spelling, punctuation, syntactic well-formedness, and usage could help distinguish whether the weak rubric correspondence observed for the Conventions research models under the researcher-defined mappings reflects the models, limitations of the explanatory feature set, limitations of the mappings, or some combination of these. Comparing alternative feature inventories would itself provide useful evidence about the dependence of post-hoc explanations on the explanatory vocabulary supplied to the surrogate.\par
Future analyses could also examine the stability of explanations across prompts, grades, and model architectures more formally. The current results suggest substantial cross-trait commonality, but a larger set of scoring conditions would permit quantitative evaluation of whether particular explanatory signals generalize or are prompt-specific.\par
Finally, global feature importance could be supplemented with response-level explanations. Global surrogate fidelity addresses whether the explanatory model reproduces score behavior across the response population, whereas local analyses could identify cases for which the surrogate and transformer diverge or for which unusual features contribute disproportionately to a prediction. Such analyses may be especially useful for model review, fairness investigations, and targeted examination of unexpected scores.\par
\section{Conclusion}
Transformer-based scoring models create a practical interpretability problem for automated writing assessment: strong agreement with human scores can coexist with limited visibility into the observable characteristics associated with model predictions. Feature-based surrogate methods can help address this problem, but their explanations should not be treated as self-validating.\par
The present study demonstrates a fidelity-first approach. Four feature-based representations were compared before substantive interpretation, and the Random Forest surrogates reproduced approximately 81\% to 95\% of DeBERTa prediction variance on held-out responses. Their SHAP profiles showed that Expression and Conventions relied on substantially overlapping observable response characteristics while differentiating the traits through relative feature weighting. Under the researcher-defined mappings, rubric correspondence was strong for Expression but weak for Conventions. Because these descriptive percentages are conditional on the mappings and available linguistic features, the Conventions result does not by itself establish a scoring-validity problem; instead, it identifies an area for further investigation of the research models and feature representation.\par
The broader implication is methodological rather than specific to any single feature ranking. Post-hoc explainability is most defensible when the explanatory model first demonstrates that it can reproduce the behavior being interpreted. Even then, the resulting explanation concerns observable score behavior rather than the internal reasoning of the transformer. Used within those boundaries, feature-based explainability can add a diagnostic layer to conventional automated-scoring validation by helping assessment developers examine apparent scoring priorities, trait differentiation, and correspondence with intended constructs.\par
\balance
\section*{References}
\noindent\hangindent=1.5em\hangafter=1 American Educational Research Association, American Psychological Association, \& National Council on Measurement in Education. (2014). \textit{Standards for educational and psychological testing}. American Educational Research Association.\par
\noindent\hangindent=1.5em\hangafter=1 Hannah, L., Jang, E. E., Shah, M., \& Gupta, V. (2023). Validity arguments for automated essay scoring of young students' writing traits. \textit{Language Assessment Quarterly, 20}(4--5), 399--420. \url{https://doi.org/10.1080/15434303.2023.2288253}\par
\noindent\hangindent=1.5em\hangafter=1 Ke, Z., \& Ng, V. (2019). Automated essay scoring: A survey of the state of the art. In \textit{Proceedings of the Twenty-Eighth International Joint Conference on Artificial Intelligence} (pp. 6300--6308). International Joint Conferences on Artificial Intelligence Organization. \url{https://doi.org/10.24963/ijcai.2019/879}\par
\noindent\hangindent=1.5em\hangafter=1 Lundberg, S. M., \& Lee, S.-I. (2017). A unified approach to interpreting model predictions. In \textit{Advances in Neural Information Processing Systems} (Vol. 30, pp. 4765--4774).\par
\noindent\hangindent=1.5em\hangafter=1 Mayfield, E., \& Black, A. W. (2020). Should you fine-tune BERT for automated essay scoring? In \textit{Proceedings of the Fifteenth Workshop on Innovative Use of NLP for Building Educational Applications} (pp. 151--162). Association for Computational Linguistics. \url{https://doi.org/10.18653/v1/2020.bea-1.15}\par
\noindent\hangindent=1.5em\hangafter=1 Nadeem, F., Nguyen, H., Liu, Y., \& Ostendorf, M. (2019). Automated essay scoring with discourse-aware neural models. In \textit{Proceedings of the Fourteenth Workshop on Innovative Use of NLP for Building Educational Applications} (pp. 484--493). Association for Computational Linguistics. \url{https://doi.org/10.18653/v1/W19-4450}\par
\noindent\hangindent=1.5em\hangafter=1 Ribeiro, M. T., Singh, S., \& Guestrin, C. (2016). ``Why should I trust you?'': Explaining the predictions of any classifier. In \textit{Proceedings of the 22nd ACM SIGKDD International }\textit{Conference on Knowledge Discovery and Data Mining} (pp. 1135--1144). Association for Computing Machinery. \url{https://doi.org/10.1145/2939672.2939778}\par
\noindent\hangindent=1.5em\hangafter=1 Schwalbe, G., \& Finzel, B. (2024). A comprehensive taxonomy for explainable artificial intelligence: A systematic survey of surveys on methods and concepts. \textit{Data Mining and Knowledge Discovery, 38}(5), 3043--3101. \url{https://doi.org/10.1007/s10618-022-00867-8}\par
\noindent\hangindent=1.5em\hangafter=1 Wang, M., Chen, Y., Huang, X., \& Lai, Y. (2026). Opening the blackbox of LLM-based automated essay scoring: Insights into feature weighting patterns and score validity. \textit{Computers and Education: Artificial Intelligence, 10}, 100568. \url{https://doi.org/10.1016/j.caeai.2026.100568}\par
\noindent\hangindent=1.5em\hangafter=1 Williamson, D. M., Xi, X., \& Breyer, F. J. (2012). A framework for evaluation and use of automated scoring. \textit{Educational Measurement: Issues and Practice, 31}(1), 2--13. \url{https://doi.org/10.1111/j.1745-3992.2011.00223.x}\par
\end{document}